%% file: authorFinal.tex
\documentclass[graybox]{SNmult}

\usepackage{type1cm}        
\usepackage{makeidx}         
\usepackage{graphicx}        
\usepackage{multicol}        
\usepackage[bottom]{footmisc}

\usepackage{newtxtext}       %
\usepackage[varvw]{newtxmath}       
\usepackage{hyperref} 

\makeindex             

\begin{document}
\title*{Evidence-Traceable Dynamic Interviewer Architecture for Expertise-Adaptive Qualitative Interviews Using Local LLMs}
\titlerunning{Evidence-Traceable Dynamic Interviewer Architecture}
\author{Aisvarya Adeseye\orcidID{0009-0003-2401-3076} and\\
Jouni Isoaho\orcidID{0000-0002-5789-3992} and\\
Adeyemi Adeseye\orcidID{0009-0006-4184-3436} and\\
Seppo Virtanen\orcidID{0000-0002-9487-3018} and\\
Mohammad Tahir\orcidID{0000-0002-6273-4603}}

\authorrunning{A. Adeseye et al.}

\institute{
Aisvarya Adeseye \at Department of Computing, University of Turku, Turku, Finland,
\email{aisvarya.a.adeseye@utu.fi}
\and
Jouni Isoaho \at Department of Computing, University of Turku, Turku, Finland,
\email{jouni.isoaho@utu.fi}
\and
Adeyemi Adeseye \at Department of Computing, University of Turku, Turku, Finland,
\email{adeyemi.a.adeseye@utu.fi}
\and
Seppo Virtanen \at Department of Computing, University of Turku, Turku, Finland,
\email{seppo.virtanen@utu.fi}
\and
Mohammad Tahir \at Department of Computing, University of Turku, Turku, Finland,
\email{tahir.mohammad@utu.fi}
}

%
%
\maketitle

\abstract{Automated interviewers and conversational agents are increasingly used in research, recruitment, customer service, and education. However, many existing systems rely on fixed question sequences and provide limited context-based personalization without considering participants’ knowledge, which can lead to repetitive or irrelevant follow-up questions.
Therefore, there is a need for an adaptive interviewing system that can adjust question depth while maintaining conversational continuity and semantic progression. 
To address this, an Evidence-Traceable Dynamic Interviewer Architecture is presented using a locally hosted Large Language Model (LLM), with the interview continuously adapted throughout the entire conversation based on the participant’s responses and evolving context.
The interviewer profiles participants' expertise in real time to generate
knowledge-appropriate questions, well-articulated responses, and smooth
transition messages that support conversational continuity.
A five-module prompt-driven architecture and persistent interview-state
record support these functions. The interviewer was evaluated with 246
participants. Expertise Profiling module (M3) showed 78.9\% exact
agreement with independently reported participant expertise, with a
weighted Cohen's \(\kappa\) of 0.80. Generate Iterative Questions module
(M4) showed a strong expertise-complexity association
(\(\rho=.79, p<.001\)), and participants reported high relevance
(mean 4.41), engagement (mean 4.32), and satisfaction (mean 4.38),
providing evidence that the architecture's adaptive components operated
consistently with their intended functions while participants reported a
positive interview experience.
}

\section{Introduction}
\label{sec:introduction}

Artificial Intelligence (AI) is increasingly used to support human--machine interaction in customer service, education, recruitment, and research~\cite{a1,a2,a3,a4}. Conversational agents and automated interviewers can reduce the effort required for repeated interactions, improve consistency, and support open-ended information elicitation~\cite{a2,a3,a4,a5,a6}. However, many existing systems still rely on predefined questions, rule-based dialogue structures, or limited intent-recognition mechanisms, restricting their ability to generate follow-up questions that respond to participants' statements, explore emerging topics, while adapting to differences in knowledge and experience~\cite{a3,a4,a5,a6}.

Large Language Models (LLMs) provide greater flexibility by interpreting conversational context and dynamically generating questions without relying entirely on hard-coded templates~\cite{a7,a8,a9,a10}. This is particularly relevant to qualitative interviewing, where participants differ in their familiarity with the topic and in the depth of discussion they can engage in. Participants with limited knowledge may require accessible, experience-based questions, whereas those demonstrating greater expertise can engage with more analytical, comparative, or evaluative questions. Natural-language responses can provide evidence for estimating evolving expertise~\cite{a11}, and recent LLM-based interviewing studies demonstrate the feasibility of adaptive, context-sensitive follow-up questioning~\cite{a7,a8,a9,a10}. However, context-sensitive generation alone does not ensure expertise-appropriate questioning or prevent semantic repetition across interview turns. LLM interaction can also produce irrelevant or unsupported outputs, inconsistent dialogue progression, hallucinations, and privacy risks~\cite{a14,a15,a16,a17}. 

Privacy is particularly important in qualitative interviewing because participant responses may contain personal, contextual, organizational, or otherwise sensitive information. Processing such responses through externally hosted LLM services may require interview data to leave the researcher's controlled computing environment. Local LLM deployment provides a
privacy-preserving alternative by enabling inference to be performed within locally controlled infrastructure and reducing the need to transmit raw participant responses to external services~\cite{a23,a7}. Previous studies have demonstrated the use of local LLMs for context-aware anonymization of sensitive qualitative data and privacy-conscious AI-powered interviewing,
while resource-efficient prompt design has shown that local deployment can remain computationally practical even under constrained hardware conditions~\cite{a23,a7,a19}.
Effective adaptive interviewing therefore requires structured mechanisms for controlling interview behavior, inferring expertise, adapting question complexity, maintaining conversational continuity, and validating generated questions. These are supported by structured prompting, controlled adaptation, output validation, and local deployment~\cite{a18,a19,a20,a21,a22,a23}.

Therefore, this study proposes an \textit{Evidence-Traceable Dynamic Interviewer Architecture} for adaptive qualitative interviewing using a locally hosted LLM. The study examines whether an AI-powered interviewer can infer participants' demonstrated expertise from their responses, use the inferred expertise to adjust subsequent question complexity, maintain contextually relevant and semantically non-repetitive questioning, and provide a positive interview experience. The architecture is designed so that the relationship between participant responses, inferred expertise, generated questions, and validation decisions can be examined throughout the interview process.

Consequently, the study addresses the following research questions:

\begin{description}

\item[\textbf{RQ1:}] How effectively can an AI-powered interviewer infer participants' expertise from their responses and use the derived expertise to adapt subsequent question complexity?

\item[\textbf{RQ2:}] How does the proposed architecture support contextually relevant and semantically non-repetitive questioning throughout an interview?

\item[\textbf{RQ3:}] How do participants evaluate the relevance, engagement, and overall experience of dynamically generated AI-powered interviews?

\end{description}

The study contributes a modular, prompt-driven, and evidence-traceable AI-powered interviewer that dynamically adjusts question complexity according to demonstrated participant expertise. It also provides a structured mechanism for generating contextually relevant, non-redundant, and semantically coherent interview dialogue using locally hosted LLM. Finally, the architecture is empirically evaluated using 246 participant interviews, examining expertise agreement, expertise-driven question adaptation, semantic question uniqueness, and participant perceptions of relevance, engagement, and overall satisfaction.

\section{Related Work}
\label{sec:related-work}

Conversational systems have increasingly been used for surveys and interviews. Studies show that conversational interfaces can improve engagement, response quality, social presence, and self-disclosure in comparison to conventional questionnaires~\cite{a2,b5,b6,b7}. Interview-specific systems have extended this work through active listening, participant profiling, and automatic follow-up questioning~\cite{a3,a4,a5,b18}. However, these approaches mainly adapt to conversational context or probing needs and rarely maintain an explicit, continuously updated representation of the participant's demonstrated expertise.

LLMs have further increased the flexibility of automated interviewing by enabling dynamic and context-sensitive question generation. Studies have shown that LLM interviewers can support scalable qualitative interviewing, generate useful probes, collect in-the-moment feedback, manage conversational goals, and maintain participant engagement~\cite{a8,a10,b8,b11,b13}. However, their performance varies across models and interview situations. Problems include inappropriate follow-ups, weak acknowledgment, limited strategic progression toward deeper questions, and variations in relevance and naturalness~\cite{a9,b14,b17}. These findings show that fluent question generation alone is insufficient; an interviewer must also control when to probe, how deeply to probe, and how questions should respond to the participant's demonstrated capability.

A key requirement for such adaptation is modeling user expertise. Research on adaptive systems shows that expertise can be inferred from observable behavior and represented as an evolving user state~\cite{b19,b20,b23,b24}. Expert--novice research further shows that experts reason using deeper principles, while novices rely more on surface characteristics~\cite{b21}. The expertise-reversal literature also indicates that interaction suitable for inexperienced users may become ineffective for more knowledgeable users~\cite{b22}. These findings support adapting question complexity according to evidence such as terminology, conceptual correctness, reasoning depth, and application or evaluative reasoning, instead of relying only on technical vocabulary.

Adaptive interviewing must also maintain conversational continuity. Active-listening research highlights the importance of acknowledging participant responses, while studies on adaptive and motivational interviewing emphasize contextual awareness, relevance, openness, and natural transitions~\cite{b14,b17}. This supports separating interviewer output into a \textbf{Response Message, Transition Message, and Next Question}, allowing acknowledgement, conversational flow, and information elicitation to be controlled as distinct but connected functions.

Another challenge is semantic repetition. Questions may use different wording while asking for essentially the same information. Semantic textual similarity methods such as Sentence-BERT, SimCSE, and sentence meta-embeddings provide mechanisms for identifying conceptual similarity beyond lexical overlap~\cite{b25,b26,b27,b28}. Although these methods are widely used for semantic similarity and duplicate detection, semantic uniqueness is less commonly implemented as an explicit \textbf{generation gate} that can reject a redundant interview question before it reaches the participant.

Prompt structure is also important for controlling LLM behavior.
Structured prompting and semantic structuring pipelines can define task instructions, persistent behavior, memory, reasoning constraints, feedback, and output formats across multi-stage workflows ~\cite{a12,b30,b31,b32,b33}. Such control is particularly important because LLM-generated outputs can contain hallucinations, contradictions, or unsupported information~\cite{a14,a16,a17}. Privacy is also important in LLM-assisted qualitative research because participant responses may contain personal, organizational, or otherwise sensitive information, creating security and privacy considerations for
AI-supported research workflows~\cite{a13,b35,b36,b37}. Altogether, these concerns support the need for structured prompts, explicit ethical and knowledge-handling rules, and locally deployed models that provide greater control over data processing.

Overall, existing research has advanced conversational surveys, adaptive probing, LLM interviewing, expertise modeling, semantic similarity, and structured prompting. However, these capabilities are generally studied separately. Expertise models describe changing user knowledge~\cite{b21,b23,b24}, while LLM interviewers mainly adapt to conversational context~\cite{b11,b13,b14}. Semantic-similarity methods identify conceptual overlap~\cite{b25,b26,b27,b28}, and structured prompting provides mechanisms for controlling LLM tasks~\cite{a24,b30,b31,b32,b33}. What remains limited is their integration into a single interviewing
process that generates semantically unique interview questions and connects participant responses to expertise estimation so that the complexity of subsequent questions is adapted while maintaining
conversational continuity.

\section{Evidence-Traceable Dynamic Interviewer Architecture}
\label{sec:3}

The Evidence-Traceable Dynamic Interviewer Architecture is designed to conduct adaptive, context-aware, and expertise-aligned qualitative interviews using a modular LLM-based pipeline. The architecture separates the main interviewing functions so that each stage can be independently configured, inspected, and evaluated. It consists of five modules: Generate System Prompt (M1), Generate Initial Question (M2), Expertise Profiling (M3), Generate Iterative Question (M4), and Validate Question Uniqueness (M5).

\begin{figure}
    \centering
    \includegraphics[width=1\linewidth]{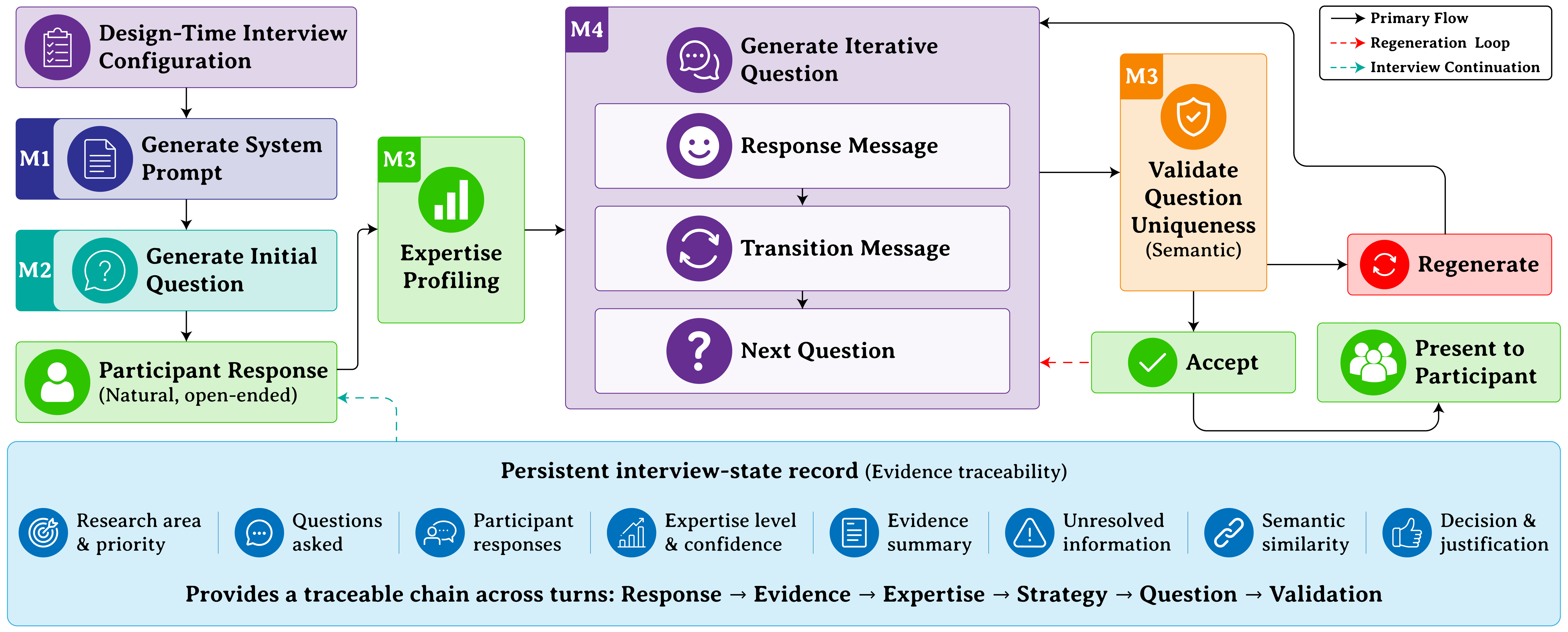}
    \caption{Overview of the Evidence-Traceable Dynamic Interviewer Architecture}
    \label{fig:overview}
\end{figure}

As illustrated in Figure~\ref{fig:overview}, the interview progresses from interview configuration and initial questioning to dynamic expertise profiling, adaptive follow-up generation, and semantic uniqueness validation. A persistent interview-state record maintains relevant information across interview turns and supports traceability between participant responses, expertise estimates, generated questions, and validation decisions. The following subsections describe each module.

\subsection{Module 1 (M1): Generate System Prompt}
\label{subsec:m1}

Module 1 establishes the interview environment by transforming the \textbf{design-time interview configuration} into a structured system prompt that governs the complete interview session. The configuration contains the research objectives, research areas and priorities, interview style, language, ethical requirements, and knowledge-handling rules. The system prompt is generated once and reused by M2--M5, providing a consistent behavioral foundation throughout the interaction.

\begin{figure}[t]
\centering
\includegraphics[width=\textwidth]{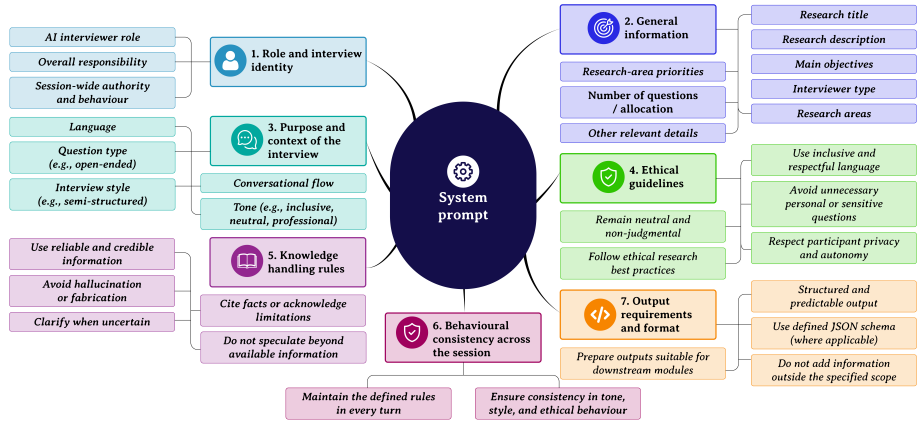}
\caption{System prompt component structure in M1. The figure presents the persistent research, behavioral, ethical, knowledge-handling, and output constraints governing the interview.}
\label{fig:m1-system-prompt}
\end{figure}

As illustrated in Figure~\ref{fig:m1-system-prompt}, the system prompt is organized into seven components: Role and Interview Identity, General Information, Purpose and Context, Ethical Guidelines, Knowledge Handling Rules, Behavioral Consistency, and Output Requirements and Format. These components separate what the interview investigates from how the interviewer communicates, handles participants, and manages factual information.

\begin{table}[t]
\centering
\caption{System-Prompt Components in M1}
\label{tab:m1-system-components}
\renewcommand{\arraystretch}{1.15}
\resizebox{\textwidth}{!}{%
\begin{tabular}{ll}
\hline
\textbf{System-Prompt Component} & \textbf{Main Function} \\
\hline
Role and Interview Identity
& Defines the AI interviewer role and responsibilities \\

General Information
& Specifies objectives, research areas, priorities, and question allocation \\

Purpose and Context
& Defines language, question type, style, tone, and conversational flow \\

Ethical Guidelines
& Controls privacy, sensitive questions, neutrality, and inclusive interaction \\

Knowledge Handling Rules
& Controls reliable information use, uncertainty, and hallucination avoidance \\

Behavioral Consistency
& Maintains the defined rules across interview turns \\

Output Requirements and Format
& Defines the structured representation used by subsequent modules \\
\hline
\end{tabular}%
}
\end{table}

These factors also support design-level traceability. Important prompt constraints can be linked to their purpose and supporting rationale. For example, restrictions on unsupported information address LLM hallucination, while privacy and sensitive-question controls support ethical handling of participant data. The evidence/source registry therefore records why important factors were included without requiring research citations to be repeatedly inserted into the interview prompts.

\begin{figure}[t]
\centering
\includegraphics[width=\textwidth]{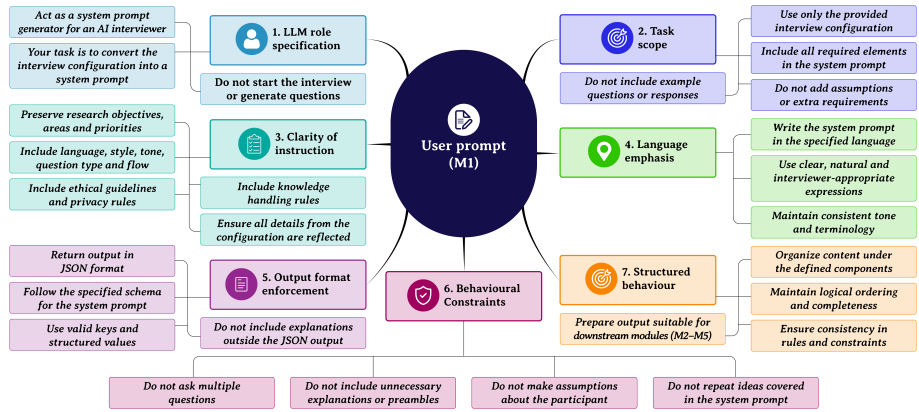}
\caption{M1 user prompt component structure. The figure presents the instruction structure used to transform the design-time interview configuration into the reusable system prompt.}
\label{fig:m1-user-prompt}
\end{figure}

M1 itself is executed through a dedicated user prompt. As illustrated in Figure~\ref{fig:m1-user-prompt}, this user prompt specifies how the LLM should convert the interview configuration into the final system prompt. It contains seven components: LLM Role Specification, Task Scope, Clarity of Instruction, Language Emphasis, Output Format Enforcement, Action Limitation, and Structured Behavior.

\begin{table}[t]
\centering
\caption{M1 User-Prompt Components}
\label{tab:m1-user-components}
\renewcommand{\arraystretch}{1.15}
\resizebox{\textwidth}{!}{%
\begin{tabular}{ll}
\hline
\textbf{M1 User-Prompt Component} & \textbf{Main Function} \\
\hline
LLM Role Specification
& Defines the model as a system-prompt generator \\

Task Scope
& Limits generation to the supplied interview configuration \\

Clarity of Instruction
& Specifies which configuration elements must be included \\

Language Emphasis
& Maintains the selected interview language and terminology \\

Output Format Enforcement
& Requires a structured, machine-readable output \\

Action Limitation
& Prevents M1 from conducting the interview or generating questions \\

Structured Behavior
& Ensures completeness and consistent organization \\
\hline
\end{tabular}%
}
\end{table}

The relationship between the two prompt structures is therefore straightforward: the M1 user prompt specifies how the system prompt should be constructed, while the resulting system prompt specifies how the interviewer should behave. Conceptually,

\begin{equation}
P_{\mathrm{sys}} = M_1\left(C,P_{\mathrm{M1}}\right),
\label{eq:m1-system-prompt}
\end{equation}

where $C$ represents the design-time interview configuration, $P_{\mathrm{M1}}$ represents the M1 user prompt, and $P_{\mathrm{sys}}$ represents the reusable system prompt passed to M2--M5.

\subsection{Module 2 (M2): Generate Initial Question}
\label{subsec:m2}

Module 2 generates the first question presented to the participant. Its purpose is to begin the interview with an accessible, open-ended, and research-relevant question that encourages a natural response while providing sufficient information for the initial expertise profiling performed by M3. M2 selects a high-priority research area and intentionally begins with a relatively simple question so that participants with different levels of knowledge can engage comfortably.

\begin{figure}[t]
    \centering
    \includegraphics[width=\columnwidth]{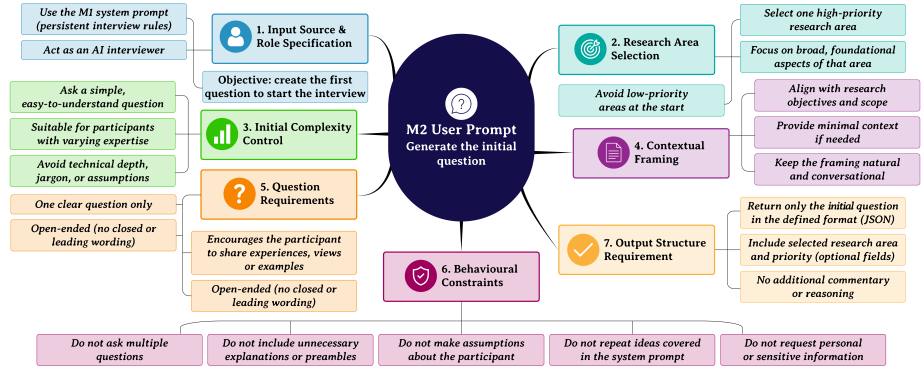}
    \caption{M2 user prompt component structure for initial question generation. The prompt defines the research-area selection, initial complexity, contextual framing, question requirements, behavioral constraints, and output structure used to generate the first interview question.}
    \label{fig:m2-user-prompt}
\end{figure}

The question is generated using the system prompt produced by M1 together with an M2-specific user prompt. The system prompt provides the overall research scope, ethical rules, language, style, and knowledge-handling requirements, while the M2 user prompt specifies how the initial question should be constructed. As illustrated in Figure~\ref{fig:m2-user-prompt}, the main elements of the M2 user prompt include research-area selection, complexity control, contextual framing, question requirements, behavioral constraints, and output structure

\begin{table}[t]
\centering
\caption{M2 User-Prompt Components for Initial Question Generation}
\label{tab:m2-components}
\begin{tabular}{p{0.38\columnwidth} p{0.54\columnwidth}}
\hline
\textbf{M2 Component} & \textbf{Function} \\
\hline
Research Area and Priority &
Selects an appropriate high-priority area for beginning the interview \\

Question Complexity &
Starts with a low-complexity, broadly understandable question \\

Contextual Framing &
Aligns the question with the research objectives and interview context \\

Question Requirements &
Produces one clear, open-ended, non-leading question \\

Behavioral Constraints &
Prevents unnecessary explanation, repetition, or multiple questions \\

Output Structure &
Returns the initial question in the required structured format \\
\hline
\end{tabular}
\end{table}

The initial question also serves as a calibration point for the adaptive interview. M2 does not determine the participant's expertise level. Instead, it generates a question that allows the participant to demonstrate their familiarity with the research topic naturally. The resulting response is then passed to M3, where observable evidence in the response is used to establish the first expertise profile.

Conceptually, M2 operates as

\begin{equation}
Q_{1} = M_{2}\left(P_{\mathrm{sys}}, R_{h}, C_{q}\right),
\label{eq:m2}
\end{equation}

where $P_{\mathrm{sys}}$ is the system prompt generated by M1, $R_{h}$ represents the selected high-priority research area, and $C_{q}$ represents the initial low-complexity questioning constraints. The output $Q_{1}$ becomes the first participant-facing question and initiates the adaptive interview cycle.

\subsection{Module 3 (M3): Expertise Profiling}
\label{subsec:m3}

Module 3 dynamically estimates the participant's demonstrated expertise from their responses and provides this information to M4 to control the depth and complexity of subsequent questions. Expertise is classified into four levels: \textit{Novice, Basic Knowledge, Advanced Knowledge, and Expert}. The classification is based on four observable factors: \textit{domain terminology and specificity (F1), conceptual correctness and relevance (F2), reasoning depth (F3), and application or evaluative reasoning (F4)}.

\begin{table*}[t]
\centering
\caption{Expertise Factors Evaluated by M3}
\label{tab:m3-expertise-factors}
\begin{tabular}{p{0.40\textwidth} p{0.62\textwidth}}
\hline
\textbf{Expertise Factor} & \textbf{What M3 Examines} \\
\hline
F1. Domain Terminology and Specificity &
Appropriate use of domain terminology, concepts, technologies, processes, or specific examples \\

F2. Conceptual Correctness and Relevance &
Whether objectively verifiable domain statements are coherent, relevant, and technically appropriate \\

F3. Reasoning Depth &
Ability to explain relationships, causes, consequences, dependencies, or trade-offs \\

F4. Application and Evaluation &
Ability to apply knowledge to situations, compare alternatives, justify decisions, or propose solutions \\
\hline
\end{tabular}
\end{table*}

M3 evaluates these factors collectively rather than relying on individual keywords or technical terms. It produces an expertise level and supporting evidence, providing a traceable basis for each classification. Personal experiences and opinions are not assessed for factual correctness; correctness is considered only when objectively verifiable domain claims are made.
The expertise profile is dynamic. After each participant response, M3 reassesses the available evidence and may maintain, increase, or decrease the current expertise estimate. The profile is represented as

\begin{equation}
E_t =
\left\{
L_t,\,
F1_t,\,
F2_t,\,
F3_t,\,
F4_t
\right\},
\label{eq:m3-expertise}
\end{equation}

where $L_t$ represents the expertise level at interview turn $t$ and $F1_t$--$F4_t$ represent the observed expertise factors.
The M3 output is stored in the persistent interview-state record and passed to M4, enabling subsequent questions to be generated at an appropriate cognitive and technical level. M3 therefore provides the adaptive link between \textit{what the participant demonstrates} and \textit{how the interviewer subsequently questions the participant}.

\begin{figure}[t]
    \centering
    \includegraphics[width=\columnwidth]{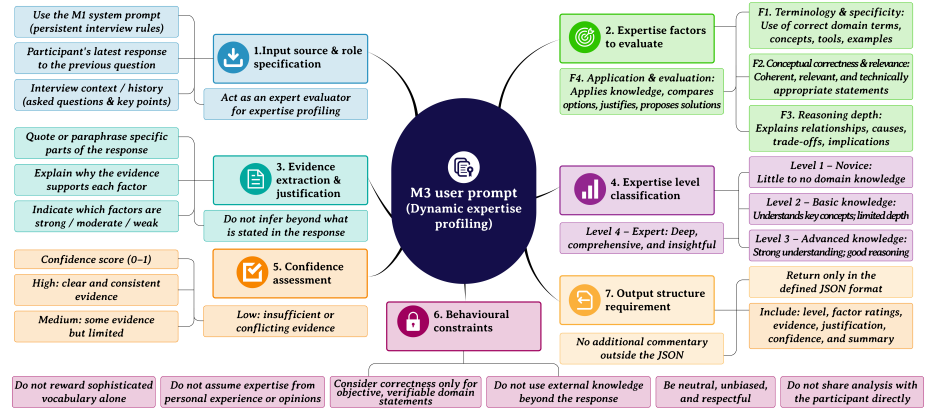}
    \caption{M3 user prompt component structure for dynamic expertise profiling. The prompt defines the participant-response input, expertise factors, classification levels, evidence requirements, behavioral constraints, and structured output used to generate the dynamic expertise profile.}
    \label{fig:m3-user-prompt}
\end{figure}

\subsection{Module 4 (M4): Generate Iterative Question}
\label{subsec:m4}

Module 4 generates the subsequent interview interaction after each participant response. It combines the participant's latest response, expertise profile from M3, previous interview context, research-area priority, and information already collected to determine an appropriate follow-up while maintaining conversational continuity and adapting question complexity to the participant's demonstrated expertise.

M4 produces three participant-facing elements: a \textit{Response Message}, a \textit{Transition Message}, and the \textit{Next Question}. The response message briefly acknowledges the participant's previous answer without introducing a new question. The transition message connects the acknowledgment to the next topic or line of inquiry, while the next question advances the interview using an open-ended question aligned with the participant's expertise and the research objectives.

\begin{table}[t]
\centering
\caption{Participant-Facing Outputs Generated by M4}
\label{tab:m4-outputs}
\begin{tabular}{p{0.33\columnwidth} p{0.59\columnwidth}}
\hline
\textbf{M4 Output} & \textbf{Function} \\
\hline
Response Message &
Briefly acknowledges or reflects the participant's previous response \\

Transition Message &
Creates a natural conversational connection between the previous response and the next question \\

Next Question &
Generates a context-aware, expertise-aligned, open-ended follow-up question \\
\hline
\end{tabular}
\end{table}

The adaptive behavior of M4 is controlled primarily by the expertise information generated by M3. Lower demonstrated expertise leads to clearer and less technically demanding questions, while higher expertise supports deeper analytical, comparative, scenario-based, or evaluative questioning. Expertise does not determine the question alone. M4 also considers what the participant has already discussed, which research areas remain insufficiently explored, and whether clarification, elaboration, examples, comparison, reflection, or deeper reasoning would best advance the interview.

The three outputs are generated as a single structured interaction so that they remain semantically connected:

\begin{equation}
O_t^{M4}
=
\left\{
R_t,\,
T_t,\,
Q_{t+1}
\right\},
\label{eq:m4-output}
\end{equation}

where $R_t$ represents the response message acknowledging the participant's latest response, $T_t$ represents the transition message, and $Q_{t+1}$ represents the proposed next question.
The candidate next question is subsequently passed to M5 for semantic uniqueness validation before presentation. If M5 identifies substantial conceptual overlap with an earlier question, the candidate is returned to M4 for regeneration.

\begin{figure}[t]
    \centering
    \includegraphics[width=\columnwidth]{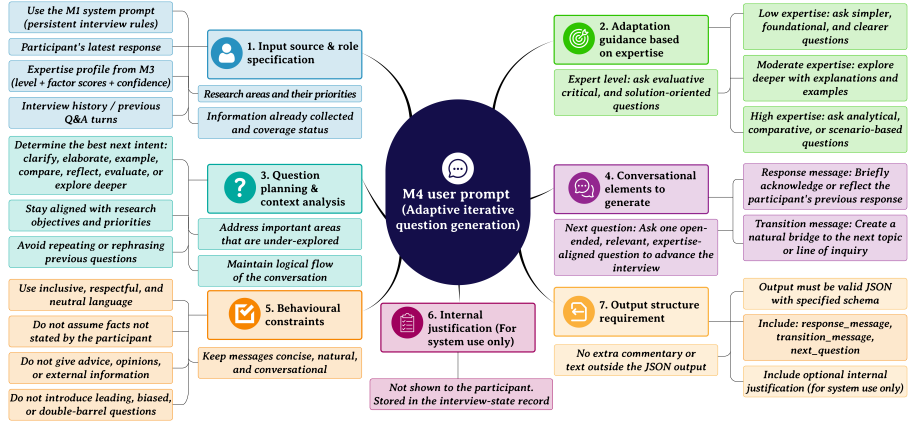}
    \caption{M4 user prompt component structure for adaptive iterative question generation. The prompt combines participant response, expertise profile, interview history, research priorities, and conversational constraints to generate the response message, transition message, and next question.}
    \label{fig:m4-user-prompt}
\end{figure}

\subsection{Module 5 (M5): Validate Question Uniqueness}
\label{subsec:m5}

Module 5 validates the \textbf{semantic uniqueness} of each candidate question generated by M4 before it is presented to the participant. Its purpose is to prevent conceptually equivalent questions from being repeated, even when they use different wording, synonymous expressions, or alternative phrasing. This supports thematic progression and reduces semantic duplication throughout the interview.
M5 compares the candidate question with the history of previously asked questions and determines whether it introduces a sufficiently distinct line of inquiry. The validation focuses on semantic overlap, contextual relevance, research progression, and duplication risk.

\begin{table}[t]
\centering
\caption{M5 Components for Semantic Question Validation}
\label{tab:m5-components}
\begin{tabular}{p{0.38\columnwidth} p{0.54\columnwidth}}
\hline
\textbf{M5 Component} & \textbf{Function} \\
\hline
Candidate Question Input &
Receives the proposed next question generated by M4 \\

Question-History Comparison &
Compares the candidate question with previously asked questions \\

Semantic Overlap Assessment &
Identifies conceptual similarity beyond exact wording \\

Research Progression Check &
Determines whether the question contributes new information or advances the discussion \\

Validation Decision &
Produces an Accept or Regenerate decision \\

Structured Output &
Returns the validation result and internal justification \\
\hline
\end{tabular}
\end{table}

The distinction between wording and meaning is important. For example, \textit{``What privacy risks do employees face when using public LLMs?''} and \textit{``What data-protection problems may occur when staff use publicly available generative AI?''} are phrased differently but seek substantially the same information and are therefore treated as semantically overlapping. In contrast, a follow-up such as \textit{``What controls would you recommend to mitigate those risks?''} advances the discussion from risk identification to mitigation and can therefore be accepted.

Conceptually, the validation can be expressed as

\begin{equation}
V_{t+1} = M_{5}\left(Q_{t+1},H_t\right),
\label{eq:m5-validation}
\end{equation}

where $Q_{t+1}$ represents the candidate next question generated by M4 and $H_t$ represents the history of questions asked up to interview turn $t$. The resulting decision $V_{t+1}$ is either Accept or Regenerate. An accepted question is presented to the participant, while a question with substantial semantic overlap is returned to M4 for regeneration.
The validation decision and its internal justification are stored in the persistent interview-state record, supporting traceability between the candidate question and the final decision.

\begin{figure}[t]
    \centering
    \includegraphics[width=\columnwidth]{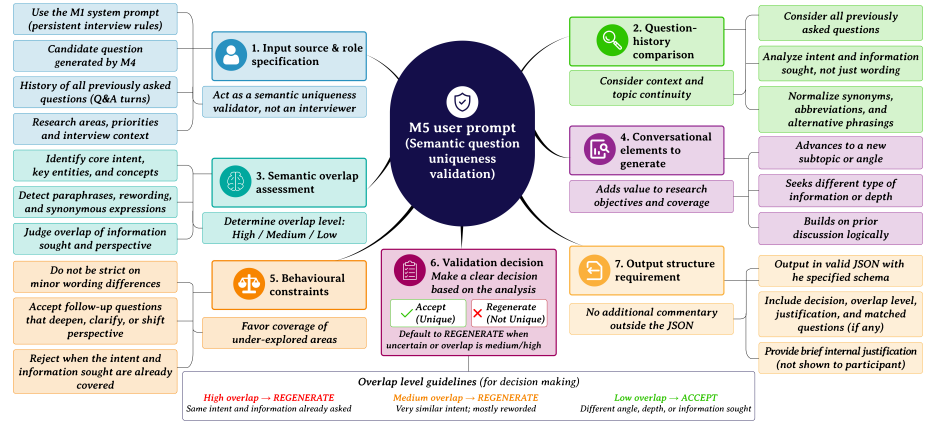}
    \caption{M5 user prompt component structure for semantic question uniqueness validation. The prompt defines the candidate-question input, question-history comparison, semantic-overlap assessment, research-progression criteria, behavioral constraints, validation decision, and structured output.}
    \label{fig:m5-user-prompt}
\end{figure}

\section{Experiment and Verification Setup}
\label{sec:experiment}

The proposed dynamic interviewer was implemented and evaluated through a workplace-oriented case study designed to examine both the operation of the five-module architecture and the quality of the resulting interview experience. The experimental setup focused on four aspects: dynamic expertise profiling, expertise-adaptive question generation, semantic question-uniqueness validation, and participant evaluation of the interview.

\subsection{Case Study and Interview Configuration}
\label{subsec:case-study}

The case study investigated \textit{how employees interact with Large Language Models (LLMs) in the workplace}. This context was selected because employees may differ substantially in their awareness, practical experience, technical understanding, and organizational use of LLMs, providing a suitable setting for evaluating dynamic expertise adaptation.
The interview covered five research areas defined for the study: awareness and knowledge of LLMs, organizational applications, skills and training, privacy and security, and organizational guidelines for LLM adoption. Each area was assigned a priority and expected question allocation, as shown in Table~\ref{tab:research-areas}.

\begin{table}[t]
\centering
\caption{Research Areas and Interview Configuration}
\label{tab:research-areas}
\begin{tabular}{p{0.59\columnwidth} p{0.18\columnwidth} p{0.13\columnwidth}}
\hline
\textbf{Research Area} & \textbf{Priority} & \textbf{Total Question Allocation} \\
\hline
Awareness and knowledge of LLMs among employees & High & 4 \\
Application of LLMs in the organization & Medium & 3 \\
Skill levels and training in using LLMs & High & 3 \\
Data privacy and security in LLM use & Medium & 4 \\
Organizational guidelines for LLM use and adoption & Low & 2 \\
\hline
\end{tabular}
\end{table}

These specifications were provided as the design-time configuration used by M1. The research areas established the expected interview coverage, while their priorities guided M2 and M4 when selecting appropriate areas for initial and subsequent questioning. The configuration specified a total of 16 participant-facing questions per interview. The first question was generated by M2 as the initial calibration question, while the remaining 15 questions were generated adaptively by M4. Accordingly, across 246 completed interviews, the analysis of adaptive follow-up questions comprised 3,690 M4-generated questions ($246 \times 15$); the initial M2 question was excluded from the follow-up-question analysis.

\subsection{Technical Implementation}
\label{subsec:technical-implementation}

The complete architecture was implemented using a locally hosted
Llama 3.2 3B model.\footnote{Meta. \textit{Llama 3.2 3B Model Card}.
Available at:
\href{https://huggingface.co/meta-llama/Llama-3.2-3B}
{https://huggingface.co/meta-llama/Llama-3.2-3B}}
The same model was used across the five modules of the modular interviewer. The same model was used across the five modules of the modular interviewer. Local execution allowed all modules to operate within the same controlled environment and avoided sending participant interview responses to an external LLM service.

Each module was implemented using the structured system and user prompts described in the preceding sections. \textbf{M1} generated the persistent system prompt from the interview configuration. \textbf{M2} generated the initial low-complexity question from a high-priority research area. After every participant response, \textbf{M3} evaluated demonstrated expertise and produced an expertise profile. \textbf{M4} used this profile together with the response and interview history to generate the response message, transition message, and next question. Finally, \textbf{M5} examined the proposed question for semantic duplication before it was presented to the participant.

The operational sequence can be represented as

\begin{equation}
M_1 \rightarrow M_2 \rightarrow R_t \rightarrow M_3
\rightarrow M_4 \rightarrow M_5 \rightarrow Q_{t+1}
\rightarrow R_{t+1},
\label{eq:experimental-flow}
\end{equation}

where $R_t$ represents the participant response at interview turn $t$, $Q_{t+1}$ represents the validated next question presented to the participant, and $R_{t+1}$ represents the participant's subsequent
response. When M5 rejected a candidate question because of semantic
overlap, the question was returned to M4 for regeneration before being
presented to the participant.
Information generated during each turn was retained in the persistent interview-state record, including previous questions, participant responses, research-area coverage, expertise estimates, and validation decisions. This provided the information required for subsequent adaptation and post-interview verification.

\subsection{Participants, Recruitment, and Experimental Procedure}
\label{subsec:participants-procedure}

Participants were recruited through university and professional networks and were required to be at least 18 years old with workplace experience. Participation was voluntary, and informed consent was obtained electronically before the interview. Participants were informed about the study purpose, data handling, confidentiality, and their right to withdraw at any time. The study was conducted online in English, and 246 completed interviews were included in the analysis. Each participant followed the same general flow, while the actual sequence and complexity of questions could vary according to their responses and dynamically estimated expertise.

M3 reassessed the participant's demonstrated expertise and assigned one of four levels: \textit{Novice}, \textit{Basic Knowledge}, \textit{Advanced Knowledge}, or \textit{Expert}.


However, for verification purposes, participants were also asked to indicate their own perceived expertise level using the same four-level classification. The self-reported expertise was collected independently from the M3 prediction and was not provided to M3 as an input during expertise profiling. This separation ensured that the adaptive interviewer relied on evidence contained in participants' responses instead of directly reproducing their self-assessment.
After the interview, participants were additionally asked to rate their interview experience. These post-interview ratings provided participant-level evidence for evaluating whether the dynamically generated interaction remained relevant, engaging, and satisfactory.

\subsection{Expertise-Profiling Verification}
\label{subsec:expertise-verification}

The first verification focused on M3 expertise profiling. Although M3 reassessed expertise after each participant response, the final expertise classification produced at the end of the interview was used for participant-level verification. This classification was compared with the participant's self-reported expertise category. This resulted in two independently obtained expertise indicators:

\begin{equation}
E_{\mathrm{LLM}} =
\text{final expertise inferred by M3},
\label{eq:llm-expertise}
\end{equation}

and

\begin{equation}
E_{\mathrm{self}} =
\text{expertise reported by the participant}.
\label{eq:self-expertise}
\end{equation}

Agreement between these classifications indicates how closely the response-based expertise profile corresponds with participants' own assessment of their knowledge.

Because the four expertise categories are ordered, verification considered both exact agreement and the magnitude of disagreement. A difference between \textit{Advanced Knowledge} and \textit{Expert}, for example, represents a smaller discrepancy than a difference between \textit{Novice} and \textit{Expert}. Percentage agreement, a confusion matrix, and linearly weighted Cohen's $\kappa$ were therefore used to evaluate agreement between the two classifications.
The participant-reported expertise level was treated as a verification reference rather than absolute ground truth, since self-assessment may overestimate or underestimate actual knowledge. The analysis therefore evaluates correspondence between two independently obtained indicators of expertise.

\subsection{Adaptive Question Verification}
\label{subsec:adaptive-question-verification}

The second verification examined whether the expertise profile generated by M3 influenced the complexity of the subsequent questions produced by M4. The expected relationship was

\begin{equation}
\text{Demonstrated Expertise}
\rightarrow
\text{Question Complexity}.
\label{eq:expertise-complexity}
\end{equation}

To evaluate this relationship, M4 follow-up questions were classified into four ordinal cognitive-complexity levels: accessible/descriptive (1), explanatory or example-based (2), analytical or comparative (3), and evaluative, scenario-based, or application-oriented (4). The classification was based on the primary cognitive demand required to answer each question. Lower levels emphasize description, experience, and straightforward explanation, while higher levels require comparison, analysis, evaluation, application, or justification.

Each generated follow-up question was therefore associated with both the M3 expertise level assigned for the corresponding interview turn and a question-complexity score from 1 to 4. For participant-level analysis, the complexity scores of the adaptive follow-up questions were aggregated to obtain a mean question-complexity score for each participant. This avoided treating multiple questions generated for the same participant as independent observations.
The relationship between M3 expertise and question complexity was evaluated using Spearman's rank correlation, since both expertise and question complexity represent ordered levels. Mean question-complexity scores were also examined across the four M3 expertise categories to determine whether question depth increased systematically from Novice to Expert.
The complexity classifications were independently assigned by two researchers during post-interview analysis using the four-level criteria defined above. Across the 246 completed interviews, a total of 3,690 adaptive follow-up questions generated by M4 were included in the complexity assessment. Inter-rater reliability was assessed using
linearly weighted Cohen's $\kappa$, since the four complexity categories were ordinal. Percentage agreement was also calculated. Disagreements were reviewed jointly by the two researchers, and a final consensus classification was assigned before the subsequent analysis.
M5 provided a separate examination of whether the adapted questions introduced a new information objective or were identified as semantically overlapping with questions already asked. Its validation history was therefore used to examine first-pass acceptance and regeneration decisions.

\subsection{Participant-Based Verification}
\label{subsec:participant-verification}

The final verification evaluates the architecture from the participant's perspective. At the end of each interview, participants rated their experience using three evaluation dimensions: Question Relevance and Coherence, Cognitive and Emotional Engagement, and Overall User Satisfaction.

\begin{table*}[t]
\centering
\caption{Participant-Based Evaluation Dimensions}
\label{tab:participant-evaluation}
\begin{tabular}{p{0.34\textwidth} p{0.58\textwidth}}
\hline
\textbf{Evaluation Dimension} & \textbf{Purpose} \\
\hline
Question Relevance and Coherence &
Evaluates whether questions were appropriate, logically connected, and relevant to the ongoing discussion \\

Cognitive and Emotional Engagement &
Evaluates whether the interaction maintained participants' attention and involvement \\

Overall User Satisfaction &
Evaluates participants' overall perception of the AI-powered interview experience \\
\hline
\end{tabular}
\end{table*}

The experimental setup provides complementary forms of verification. Architecture-level verification examines the correspondence between M3 expertise profiles and participant-reported expertise, the relationship between M3 expertise and subsequent M4 question complexity, and the semantic-overlap decisions produced by M5. Participant-level verification examines whether the resulting interviews are perceived as relevant, engaging, and satisfactory.

\begin{table*}[t]
\centering
\caption{Summary of Experimental Verification Components}
\label{tab:verification-components}
\begin{tabular}{p{0.27\textwidth} p{0.35\textwidth} p{0.30\textwidth}}
\hline
\textbf{Verification Component} &
\textbf{Comparison / Evidence} &
\textbf{Main Purpose} \\
\hline

Expertise Profiling &
M3 expertise vs. participant-reported expertise &
Assess agreement in expertise classification \\

Question Adaptation &
M3 expertise vs. subsequent M4 question complexity &
Assess expertise-driven adaptation \\

Question Uniqueness &
M5 decisions and question-history comparisons &
Examine semantic-overlap filtering and regeneration \\

Question Relevance and Coherence &
Participant post-interview rating &
Assess conversational quality \\

Engagement &
Participant post-interview rating &
Assess participant involvement \\

Overall Satisfaction &
Participant post-interview rating &
Assess overall interview experience \\
\hline
\end{tabular}
\end{table*}

Together, these measures evaluate both the internal adaptive behavior of the architecture and the participant experience produced by the resulting interviews.

\section{Results}
\label{sec:results}

The results are presented across the main components evaluated in the experiment. The demographic and background characteristics of the participants are first described, followed by the verification of M3 expertise profiling, M4 expertise-adaptive question generation, and M5 semantic question-uniqueness validation. Representative interview questions are included to illustrate how question depth changed across expertise levels. Finally, participants' evaluations of the overall interview experience are reported.

\subsection{Participant Characteristics}
\label{subsec:participant-characteristics}

A total of 246 participants completed the AI-powered interview. The sample included participants from different age groups, educational backgrounds, employment sectors, levels of professional experience, and frequencies of LLM use. Table~\ref{tab:participant-characteristics} presents the demographic and background characteristics of the participants.

\begin{table*}[t]
\centering
\caption{Participant Demographic and Background Characteristics ($N=246$)}
\label{tab:participant-characteristics}
\small
\renewcommand{\arraystretch}{1.08}
\setlength{\tabcolsep}{10pt}

\begin{tabular}{l r@{\hspace{2cm}} l r}
\hline
\textbf{Characteristic} & \textbf{$n$ (\%)} &
\textbf{Characteristic} & \textbf{$n$ (\%)} \\
\hline

\multicolumn{2}{l}{\textbf{Gender}} &
\multicolumn{2}{l}{\textbf{Age}} \\

Male & 130 (52.8) &
18--24 years & 31 (12.6) \\

Female & 111 (45.1) &
25--34 years & 78 (31.7) \\

Prefer not to say & 5 (2.0) &
35--44 years & 70 (28.5) \\

& &
45--54 years & 43 (17.5) \\

& &
55 years or above & 24 (9.8) \\[2pt]

\multicolumn{2}{l}{\textbf{Highest education}} &
\multicolumn{2}{l}{\textbf{Work experience}} \\

Upper-secondary or vocational & 32 (13.0) &
Less than 2 years & 36 (14.6) \\

Bachelor's degree & 91 (37.0) &
2--5 years & 64 (26.0) \\

Master's degree & 99 (40.2) &
6--10 years & 60 (24.4) \\

Doctorate & 18 (7.3) &
11--15 years & 42 (17.1) \\

Other & 6 (2.4) &
More than 15 years & 44 (17.9) \\[2pt]

\multicolumn{2}{l}{\textbf{Employment sector}} &
\multicolumn{2}{l}{\textbf{Frequency of LLM use}} \\

Technology/IT & 61 (24.8) &
Daily & 72 (29.3) \\

Professional services & 46 (18.7) &
Weekly & 86 (35.0) \\

Education/research & 38 (15.4) &
Monthly & 41 (16.7) \\

Public sector & 31 (12.6) &
Rarely & 31 (12.6) \\

Healthcare & 24 (9.8) &
Never & 16 (6.5) \\

Finance & 19 (7.7) & & \\
Manufacturing/engineering & 17 (6.9) & & \\
Other & 10 (4.1) & & \\

\hline
\end{tabular}
\end{table*}

The participant characteristics show considerable variation in professional experience and familiarity with LLM technologies. Overall, 64.2\% of participants reported using LLMs either daily or weekly, whereas 19.1\% reported rarely or never using them. This variation provided a suitable participant population for evaluating an architecture designed to adjust interview questioning according to different levels of demonstrated expertise.
Participants also independently reported their perceived level of LLM expertise using the same four categories applied by M3. Based on the participant-reported classifications, 52 participants (21.1\%) identified as Novice, 73 (29.7\%) as Basic Knowledge, 76 (30.9\%) as Advanced Knowledge, and 45 (18.3\%) as Expert. These values were collected independently and were not provided to M3 during expertise profiling.

\subsection{Expertise-Profiling Results}
\label{subsec:expertise-profiling-results}

The first architecture-level verification examined the agreement between the expertise level inferred by M3 and the expertise level independently reported by each participant. Table~\ref{tab:expertise-confusion} presents the resulting classification matrix.

\begin{table}[t]
\centering
\caption{M3 Expertise Classification Compared with Participant Self-Assessment}
\label{tab:expertise-confusion}
\renewcommand{\arraystretch}{1.10}

\begin{tabular}{lcccc}
\hline
\textbf{Self-Reported Expertise}
& \textbf{M3 Novice}
& \textbf{M3 Basic}
& \textbf{M3 Advanced}
& \textbf{M3 Expert} \\
\hline
Novice
& \textbf{44} & 7 & 1 & 0 \\

Basic Knowledge
& 8 & \textbf{54} & 10 & 1 \\

Advanced Knowledge
& 1 & 9 & \textbf{58} & 8 \\

Expert
& 0 & 1 & 6 & \textbf{38} \\
\hline
\end{tabular}

\end{table}

M3 assigned the same expertise category as the participant's self-assessment for \textit{194 of the 246 participants}, corresponding to an \textit{exact agreement of 78.9\%}. When classifications differing by only one adjacent expertise level were considered, agreement increased to \textit{98.4\%}. A linearly weighted Cohen's $\kappa$ of \textit{0.80} further indicated strong agreement between the two ordinal classifications.
Most disagreements occurred between neighboring categories. Participants reporting Basic Knowledge were occasionally classified as Advanced Knowledge, while some Advanced Knowledge participants were classified as either Basic Knowledge or Expert. Only four participants differed by more than one expertise category. These findings indicate that the response-based expertise profile generated by M3 generally corresponded with participants' independently reported expertise while remaining independent of the self-assessment itself.
Participant-reported expertise was used as an independent verification reference rather than an absolute ground truth, since self-assessment may itself involve over- or underestimation of knowledge.

\subsection{Expertise-Adaptive Question Generation}
\label{subsec:adaptive-question-results}

The next evaluation examined whether the expertise levels inferred by M3 were reflected in the questions subsequently generated by M4. Follow-up questions were classified into four ordinal cognitive-complexity levels: accessible/descriptive (1), explanatory or example-based (2), analytical or comparative (3), and evaluative, scenario-based, or application-oriented (4).
Inter-rater reliability for the question-complexity classifications was high. The two researchers achieved 89.1\% exact agreement across the 3,690 independently classified follow-up questions, with a linearly weighted Cohen's $\kappa$ of 0.86. Most disagreements involved adjacent complexity levels and were resolved through discussion before the final
classifications were used in the subsequent analysis.

\begin{table}[t]
\centering
\caption{Question Complexity by M3 Expertise Level}
\label{tab:question-complexity}
\renewcommand{\arraystretch}{1.10}
\begin{tabular}{lcc}
\hline
\textbf{M3 Expertise Level} & \textbf{Mean Complexity} & \textbf{SD} \\
\hline
Novice             & 1.42 & 0.55 \\
Basic Knowledge    & 2.08 & 0.60 \\
Advanced Knowledge & 2.91 & 0.63 \\
Expert             & 3.56 & 0.51 \\
\hline
\end{tabular}
\end{table}

A clear increase in question complexity was observed with increasing expertise. Questions generated for Novice participants primarily requested descriptions, experiences, or straightforward examples. Basic Knowledge participants received more explanatory and application-oriented questions. Advanced Knowledge participants were increasingly asked to analyze relationships, compare alternatives, and consider risks, while Expert participants more frequently received scenario-based and evaluative questions requiring justification, assessment of trade-offs, or proposals for technical and organizational controls.

The relationship between M3 expertise level and subsequent question complexity was strong and statistically significant, Spearman's $\rho=.79$, $p<.001$. This indicates that M3 did not function only as an expertise classifier; its output systematically influenced the cognitive depth of the questions generated by M4.

\subsection{Examples of Adaptive Question Progression}
\label{subsec:adaptive-question-examples}

Representative questions illustrate how M4 adjusted the depth of questioning while remaining within the same general research area. Table~\ref{tab:adaptive-question-examples} shows the progression across the four expertise levels.

\begin{table*}[t]
\centering
\caption{Representative Questions Across Expertise Levels}
\label{tab:adaptive-question-examples}
\renewcommand{\arraystretch}{1.25}
\resizebox{\textwidth}{!}{%
\begin{tabular}{lp{12.5cm}}
\hline
\textbf{Expertise Level} & \textbf{Representative Question} \\
\hline

\textbf{Novice} &
\textit{``Have you ever thought about privacy when using an LLM at work? What kinds of information would you avoid sharing with it?''} \\

\textbf{Basic Knowledge} &
\textit{``What privacy or security problems do you think could arise when employees enter work-related information into publicly available LLMs?''} \\

\textbf{Advanced Knowledge} &
\textit{``How would you assess the risk of sensitive organizational information being exposed through employees' use of public LLM services?''} \\

\textbf{Expert} &
\textit{``Suppose an organization wants to permit employees to use public LLMs while protecting confidential information. What technical and governance controls would you recommend, and how would you evaluate whether those controls are effective?''} \\

\hline
\end{tabular}%
}
\end{table*}

The progression is expressed primarily through the depth of reasoning required, rather than simply through increased use of technical terminology. The Novice question allows participants to respond using awareness and personal experience. The Basic Knowledge question asks participants to identify and explain potential problems. The Advanced Knowledge question requires assessment of organizational risk, while the Expert question requires participants to propose and evaluate technical and governance measures.

M4 also maintained the three-part interaction consisting of a Response Message, Transition Message, and Next Question. One representative sequence was:

\begin{quote}

\noindent
\textbf{Participant Response:}

\textit{``We sometimes use public LLMs for rewriting documents, but we are told not to include client information because we do not know how that information may be processed.''}

\vspace{0.5em}

\noindent
\textbf{Response Message:}

\textit{``That highlights an important distinction between using the tool for general assistance and deciding what information can safely be shared.''}

\vspace{0.5em}

\noindent
\textbf{Transition Message:}

\textit{``The way organizations manage that boundary can involve both employee practices and formal safeguards.''}

\vspace{0.5em}

\noindent
\textbf{Next Question:}

\textit{``What controls or organizational guidelines do you think would be most effective for reducing the risk of employees exposing sensitive information through LLM use?''}

\end{quote}

The sequence illustrates how the generated response acknowledges the participant's contribution, the transition maintains conversational continuity, and the subsequent question advances the discussion from \textbf{recognition of a privacy risk to possible mitigation measures}.

\subsection{Semantic Question-Uniqueness Validation}
\label{subsec:semantic-uniqueness-results}

The operation of M5 was evaluated by examining whether candidate follow-up questions generated by M4 were semantically distinct from questions already asked within the same interview. Across the 3,690 first-pass candidate follow-up questions, M5 accepted \textit{3,469} (\textit{94.0\%}) and identified
\textit{221} (\textit{6.0\%}) as having substantial semantic overlap,
returning these candidates to M4 for regeneration.
Most rejected questions did not contain identical wording. Instead, they requested substantially the same information using alternative terminology or phrasing. For example:

\begin{quote}

\textit{``What privacy risks do employees face when using public LLMs?''}

\vspace{0.4em}

\noindent
and

\vspace{0.4em}

\textit{``What data-protection problems may occur when staff use publicly available generative AI?''}

\end{quote}

These questions were treated as semantically overlapping because both target essentially the same information. A regenerated question such as

\begin{quote}
\textit{``What controls would you recommend to mitigate those risks?''}
\end{quote}

was accepted because it extends the discussion from risk identification to mitigation, thereby adding a new information objective rather than repeating the previous question. Of the \textit{221} questions returned to M4, \textit{207} (\textit{93.7\%}) were accepted following the first regeneration cycle, while \textit{14} (\textit{6.3\%}) required two or more regeneration cycles.

\subsection{Participant Evaluation}
\label{subsec:participant-evaluation-results}
After the completion of the interview, participants evaluated their experience on a five-point scale. Three dimensions were examined: Question Relevance and Coherence, Cognitive and Emotional Engagement, and Overall User Satisfaction.

\begin{table}[t]
\centering
\caption{Participant Evaluation of the AI-Powered Interview}
\label{tab:participant-evaluation-results}
\renewcommand{\arraystretch}{1.15}

\begin{tabular}{lcc}
\hline
\textbf{Evaluation Dimension} & \textbf{Mean} & \textbf{SD} \\
\hline
Question Relevance and Coherence
    & \textbf{4.41} & \textbf{0.58} \\

Cognitive and Emotional Engagement
    & \textbf{4.32} & \textbf{0.64} \\

Overall User Satisfaction
    & \textbf{4.38} & \textbf{0.61} \\
\hline
\end{tabular}

\end{table}

Question relevance and coherence received the highest mean score ($M=4.41$, $SD=0.58$), indicating that participants generally perceived the dynamically generated questions as relevant to their responses and logically connected throughout the interview. Cognitive and emotional engagement was also rated highly ($M=4.32$, $SD=0.64$), suggesting that the adaptive questioning maintained participant involvement. Overall satisfaction was similarly positive ($M=4.38$, $SD=0.61$).
Together with the architecture-level verification, these participant ratings indicate that dynamically adapting question complexity and maintaining semantic progression did not negatively affect the interview experience. Participants generally experienced the interviews as relevant, engaging, and satisfactory while the interviewer adjusted its questioning according to their demonstrated expertise.

\section{Conclusion}
\label{sec:conclusion}

This study presented an Evidence-Traceable Dynamic Interviewer Architecture for conducting adaptive qualitative interviews using a locally hosted LLM. The architecture separates the interviewing process into five modules covering system-prompt generation, initial questioning, dynamic expertise profiling, adaptive iterative questioning, and semantic question-uniqueness validation. A persistent interview-state record supports traceability between participant responses, expertise inference, question-generation decisions, and validation outcomes.
The architecture was implemented using Llama 3.2 3B and evaluated with 246 participants in a workplace LLM-use case study. The results provide evidence that the evaluated adaptive components operated consistently with their intended functions. M3 achieved 78.9\% exact agreement with independently reported participant expertise and 98.4\% agreement within one adjacent level, with a linearly weighted Cohen's $\kappa$ of 0.80. M4 showed a strong relationship between inferred expertise and subsequent question complexity ($\rho=.79$, $p<.001$). M5 accepted 94.0\% of first-pass candidate questions and returned 6.0\% for regeneration after identifying semantic overlap. Participants also rated question relevance and coherence ($M=4.41$), engagement ($M=4.32$), and overall satisfaction ($M=4.38$) positively.

These findings show that expertise profiling serves a functional role beyond participant classification by influencing the depth and cognitive demand of subsequent questions. M5 further provides a filtering mechanism for identifying potentially repetitive questions before they are presented to participants. Together, response evidence, expertise-aware adaptation, conversational continuity, and semantic validation provide a structured process for moving from a participant's response to an appropriate next question while retaining an inspectable decision trail.

A limitation of the present study is that the architecture was evaluated using a single local LLM, Llama 3.2 3B. The architecture is conceptually model-independent because its modules and information flow are separated from the underlying LLM; however, the present evaluation does not establish model-independent performance. Different local models may respond differently to the same structured prompts, and cross-model evaluation is therefore required to determine the consistency and robustness of the architecture across model families and parameter sizes.

Future work will evaluate the architecture using multiple local LLMs to examine model independence, robustness, and consistency of expertise classification and adaptive questioning. Further studies can apply the architecture to different qualitative research domains, languages, and participant populations. Additional work may investigate longer-term interview memory and integration with downstream qualitative-analysis pipelines, supporting a more complete privacy-conscious workflow for locally executed interview collection and analysis. The modular and evidence-traceable structure presented in this study provides a foundation for developing context-aware, expertise-aligned, non-repetitive, and auditable qualitative interviews.

\section*{Ethics Approval and Informed Consent}

The study was conducted in accordance with the ethical principles for
research involving human participants issued by the Finnish National
Board on Research Integrity (TENK, 2019). Under Finnish ethical review
practice, prior ethical review is required when a study contains one or
more of the specific research-design elements defined by TENK. These
criteria are listed in Section 4.2, ``Research design elements requiring
ethical review,'' on page 19 of the TENK guidelines.\footnote{Finnish
National Board on Research Integrity (TENK). \textit{The Ethical
Principles of Research with Human Participants and Ethical Review in the
Human Sciences in Finland: TENK Guidelines 2019}, Section 4.2, p.~21 (if PDF).
Available at:
\href{https://tenk.fi/en/advice-and-materials/guidelines-ethical-review-human-sciences}
{Official TENK guidelines}}
The present study did not contain any of these research-design elements;
therefore, prior review by the Ethics Committee for Human Sciences at
the University of Turku was not required. Participation was voluntary,
and electronic informed consent was obtained from all participants
before participation.

\input{references}

\end{document}

%% file: references.tex
%
%
%
%

%% file: authorFinal.bbl
\begin{thebibliography}{99.}%
%
%

\bibitem{a1} Følstad A, Araujo T, Law ELC, Brandtzaeg PB, Papadopoulos S, Reis L, Baez M, Laban G, McAllister P, Ischen C, Wald R, Catania F, Meyer von Wolff R, Hobert S, Luger E (2021) Future directions for chatbot research: an interdisciplinary research agenda. Computing 103(12):2915--2942. doi: 10.1007/s00607-021-01016-7


\bibitem{a2} Xiao Z, Zhou MX, Liao QV, Mark G, Chi C, Chen W, Yang H (2020) Tell me about yourself: using an AI-powered chatbot to conduct conversational surveys with open-ended questions. ACM Trans Comput-Hum Interact 27(3):15. doi: 10.1145/3381804



\bibitem{a3} Xiao Z, Zhou MX, Chen W, Yang H, Chi C (2020) If I hear you correctly: building and evaluating interview chatbots with active listening skills. In: Proceedings of the 2020 CHI Conference on Human Factors in Computing Systems (CHI '20), pp 1--14. Association for Computing Machinery, New York. doi: 10.1145/3313831.3376131


\bibitem{a4} Hu J, Guo J, Tang N, Ma X, Yao Y, Yang C, Xu Y (2024) Designing the conversational agent: asking follow-up questions for information elicitation. Proc ACM Hum-Comput Interact 8(CSCW1):43. doi: 10.1145/3637320

\bibitem{a5} Han X, Zhou MX, Turner MJ, Yeh T (2021) Designing effective interview chatbots: automatic chatbot profiling and design suggestion generation for chatbot debugging. In: Proceedings of the 2021 CHI Conference on Human Factors in Computing Systems. Association for Computing Machinery, New York. doi: 10.1145/3411764.3445569

\bibitem{a6} Langer M, König CJ, Sanchez DRP, Samadi S (2020) Highly automated interviews: applicant reactions and the organizational context. J Manag Psychol 35(4):301--314. doi: 10.1108/JMP-09-2018-0402

\bibitem{a7} Adeseye A, Isoaho J, Virtanen S, Tahir M (2025) Modular AI-powered interviewer with dynamic question generation and expertise profiling. In: World Congress in Computer Science, Computer Engineering \& Applied Computing, pp 59--73. Springer Nature Switzerland, Cham

\bibitem{a8} Wuttke A, A{\ss}enmacher M, Klamm C, Lang MM, W{\"u}rschinger Q, Kreuter F (2025) AI conversational interviewing: transforming surveys with LLMs as adaptive interviewers. In: Kazantseva A, Szpakowicz S, Degaetano-Ortlieb S, Bizzoni Y, Pagel J (eds) Proceedings of the 9th Joint SIGHUM Workshop on Computational Linguistics for Cultural Heritage, Social Sciences, Humanities and Literature (LaTeCH-CLfL 2025), pp 179--204. Association for Computational Linguistics, Albuquerque. doi: 10.18653/v1/2025.latechclfl-1.17

\bibitem{a9} Panfilova A, Bolshev V, Mozikov M, Latynov V, Vanin A, Nestik T, Nourkova V, Vlasova A, Kozin M, Serohvostov A, Tarasova E, Nikolenko S (2026) The AI interviewer: multi-faceted evaluation of adaptive questioning by large language models. Sci Rep 16(1):20257. doi: 10.1038/s41598-026-46517-7

\bibitem{a10} Liu M, Wang T, Cohen CA, Li S, Xiong C (2025) Understand user opinions of large language models via LLM-powered in-the-moment user experience interviews. In: Che W, Nabende J, Shutova E, Pilehvar MT (eds) Findings of the Association for Computational Linguistics: ACL 2025, pp 13872--13893. Association for Computational Linguistics, Vienna. doi: 10.18653/v1/2025.findings-acl.714

\bibitem{a11} Adeseye A, Isoaho J, Virtanen S, Tahir M (2026) Dynamic agentic AI expert profiler system architecture for multidomain intelligence modeling. In: 2026 IEEE Conference on Artificial Intelligence (CAI), pp 1593--1598. IEEE. doi: 10.1109/CAI68641.2026.11536612

\bibitem{a12} Adeseye A, Isoaho J, Virtanen S, Tahir M (2026) A resource-efficient codebook-driven semantic structuring pipeline for human-AI dialogue in ambient intelligent systems. Procedia Comput Sci 280:552--559. doi: 10.1016/j.procs.2026.04.070

\bibitem{a13} Adeseye A, Isoaho J, Tahir M (2025) LLM-assisted qualitative data analysis: security and privacy concerns in gamified workforce studies. Procedia Comput Sci 257:60--67. doi: 10.1016/j.procs.2025.03.011

\bibitem{a14} Ji Z, Lee N, Frieske R, Yu T, Su D, Xu Y, Ishii E, Bang YJ, Madotto A, Fung P (2023) Survey of hallucination in natural language generation. ACM Comput Surv 55(12):248. doi: 10.1145/3571730

\bibitem{a15} Chen K, Zhou X, Lin Y, Feng S, Shen L, Wu P (2025) A survey on privacy risks and protection in large language models. J King Saud Univ Comput Inf Sci 37(7):163. doi: 10.1007/s44443-025-00177-1

\bibitem{a16} Schroeder H, Aubin Le Qu{\'e}r{\'e} M, Randazzo C, Mimno D, Schoenebeck S (2025) Large language models in qualitative research: uses, tensions, and intentions. In: Proceedings of the 2025 CHI Conference on Human Factors in Computing Systems (CHI '25), Article 481, pp 1--17. Association for Computing Machinery, New York. doi: 10.1145/3706598.3713120

\bibitem{a17} Roberts J, Baker M, Andrew J (2024) Artificial intelligence and qualitative research: the promise and perils of large language model (LLM) ‘assistance’. Crit Perspect Account 99:102722. doi: 10.1016/j.cpa.2024.102722

\bibitem{a18} Adeseye A, Isoaho J, Tahir M (2025) Systematic prompt framework for qualitative data analysis: designing system and user prompts. In: 2025 IEEE 5th International Conference on Human-Machine Systems (ICHMS), pp 229--234. IEEE. doi: 10.1109/ICHMS65439.2025.11154183

\bibitem{a19} Adeseye A, Isoaho J, Virtanen S, Tahir M (2025) Efficient prompt design for resource-constrained deployment of local LLMs. In: 2025 IEEE Nordic Circuits and Systems Conference (NorCAS), pp 1--7. IEEE. doi: 10.1109/NorCAS66540.2025.11231309

\bibitem{a20} Adeseye A, Isoaho J, Virtanen S, Tahir M (2025) Why compromise privacy? Local LLMs rival commercial LLMs in qualitative analysis. In: 2025 Computing, Communications and IoT Applications (ComComAp), pp 127--132. IEEE. doi: 10.1109/ComComAp68359.2025.11353130

\bibitem{a21} Adeseye A, Isoaho J, Tahir M (2026) Performance evaluation of LLM hallucination reduction strategies for reliable qualitative analysis. In: Arabnia HR, Deligiannidis L, Amirian S, Ghareh Mohammadi F, Shenavarmasouleh F (eds) AI Revolution: Research, Ethics and Society, pp 142--156. Springer Nature Switzerland, Cham. doi: 10.1007/978-3-032-12313-8\_11

\bibitem{a22} Adeseye A, Isoaho J, Virtanen S, Tahir M (2026) LLM-assisted codebook development for cybersecurity interviews with enhanced accuracy and reduced hallucination. In: 2026 IEEE 5th International Conference on AI in Cybersecurity (ICAIC), pp 1--6. IEEE. doi: 10.1109/ICAIC67076.2026.11395872

\bibitem{a23} Adeseye A, Isoaho J, Virtanen S, Tahir M (2026) Local language models for context-aware adaptive anonymization of sensitive text. In: Ferens K, Deligiannidis L, Arabnia HR, de la Fuente D, Olivas JA (eds) Applied Cognitive Computing and Artificial Intelligence, pp 190--204. Springer Nature Switzerland, Cham. doi: 10.1007/978-3-032-22205-3\_14

\bibitem{a24} Liu P, Yuan W, Fu J, Jiang Z, Hayashi H, Neubig G (2023) Pre-train, prompt, and predict: a systematic survey of prompting methods in natural language processing. ACM Comput Surv 55(9):195. doi: 10.1145/3560815



\bibitem{b5} Celino I, Re Calegari G (2020) Submitting surveys via a conversational interface: an evaluation of user acceptance and approach effectiveness. Int J Hum-Comput Stud 139:102410. doi: 10.1016/j.ijhcs.2020.102410

\bibitem{b6} Rhim J, Kwak M, Gong Y, Gweon G (2022) Application of humanization to survey chatbots: change in chatbot perception, interaction experience, and survey data quality. Comput Hum Behav 126:107034. doi: 10.1016/j.chb.2021.107034

\bibitem{b7} Zarouali B, Araujo T, Ohme J, de Vreese C (2024) Comparing chatbots and online surveys for (longitudinal) data collection: an investigation of response characteristics, data quality, and user evaluation. Commun Methods Meas 18(1):72--91. doi: 10.1080/19312458.2022.2156489

\bibitem{b8} Jacobsen RM, Cox SR, Griggio CF, van Berkel N (2025) Chatbots for data collection in surveys: a comparison of four theory-based interview probes. In: Proceedings of the 2025 CHI Conference on Human Factors in Computing Systems (CHI '25), Article 228, pp 1--21. Association for Computing Machinery, New York. doi: 10.1145/3706598.3714128



\bibitem{b11} Jack T, Cooper A, Flower L (2026) Automating the qualitative interview? Using Gen AI chatbots in social science research. Methodol Innov 20597991261448157. doi: 10.1177/20597991261448157



\bibitem{b13} Duan J, Zhao X, Zhang Z, Ko EG, Boddy L, Wang C, Li T, Rasgon A, Hong J, Lee MK, Yuan C, Long Q, Ding Y, Chen T, Xu K (2025) GuideLLM: exploring LLM-guided conversation with applications in autobiography interviewing. In: Chiruzzo L, Ritter A, Wang L (eds) Proceedings of the 2025 Conference of the Nations of the Americas Chapter of the Association for Computational Linguistics: Human Language Technologies (Volume 1: Long Papers), pp 5558--5588. Association for Computational Linguistics, Albuquerque. doi: 10.18653/v1/2025.naacl-long.287

\bibitem{b14} Spangher A, Lu M, Kalyan S, Cho HJ, Huang T, Shi W, May J (2025) NewsInterview: a dataset and a playground to evaluate LLMs' grounding gap via informational interviews. In: Che W, Nabende J, Shutova E, Pilehvar MT (eds) Proceedings of the 63rd Annual Meeting of the Association for Computational Linguistics (Volume 1: Long Papers), pp 32895--32925. Association for Computational Linguistics, Vienna. doi: 10.18653/v1/2025.acl-long.1580



\bibitem{b17} Basar E, Sun X, Hendrickx I, de Wit J, Bosse T, De Bruijn GJ, Bosch JA, Krahmer E (2025) How well can large language models reflect? A human evaluation of LLM-generated reflections for motivational interviewing dialogues. In: Rambow O, Wanner L, Apidianaki M, Al-Khalifa H, Di Eugenio B, Schockaert S (eds) Proceedings of the 31st International Conference on Computational Linguistics, pp 1964--1982. Association for Computational Linguistics, Abu Dhabi

\bibitem{b18} Xiao Z, Zhou MX, Fu WT (2019) Who should be my teammates: using a conversational agent to understand individuals and help teaming. In: Proceedings of the 24th International Conference on Intelligent User Interfaces (IUI '19), pp 437--447. Association for Computing Machinery, New York. doi: 10.1145/3301275.3302264

\bibitem{b19} Vaubel KP, Gettys CF (1990) Inferring user expertise for adaptive interfaces. Hum-Comput Interact 5(1):95--117. doi: 10.1207/s15327051hci0501\_3

\bibitem{b20} Sedera D, Dey S (2013) User expertise in contemporary information systems: conceptualization, measurement and application. Inf Manag 50(8):621--637. doi: 10.1016/j.im.2013.07.004


\bibitem{b21} Chi MTH, Feltovich PJ, Glaser R (1981) Categorization and representation of physics problems by experts and novices. Cogn Sci 5(2):121--152. doi: 10.1207/s15516709cog0502\_2


\bibitem{b22} Kalyuga S, Ayres P, Chandler P, Sweller J (2003) The expertise reversal effect. Educ Psychol 38(1):23--31. doi: 10.1207/S15326985EP3801\_4

\bibitem{b23} Brusilovsky P (2001) Adaptive hypermedia. User Model User-Adapt Interact 11(1):87--110. doi: 10.1023/A:1011143116306

\bibitem{b24} Abdelrahman G, Wang Q, Nunes B (2023) Knowledge tracing: a survey. ACM Comput Surv 55(11):224. doi: 10.1145/3569576

\bibitem{b25} Reimers N, Gurevych I (2019) Sentence-BERT: sentence embeddings using Siamese BERT-networks. In: Inui K, Jiang J, Ng V, Wan X (eds) Proceedings of the 2019 Conference on Empirical Methods in Natural Language Processing and the 9th International Joint Conference on Natural Language Processing (EMNLP-IJCNLP), pp 3982--3992. Association for Computational Linguistics, Hong Kong. doi: 10.18653/v1/D19-1410

\bibitem{b26} Gao T, Yao X, Chen D (2021) SimCSE: simple contrastive learning of sentence embeddings. In: Moens MF, Huang X, Specia L, Yih SW (eds) Proceedings of the 2021 Conference on Empirical Methods in Natural Language Processing, pp 6894--6910. Association for Computational Linguistics, Online and Punta Cana. doi: 10.18653/v1/2021.emnlp-main.552

\bibitem{b27} Poerner N, Waltinger U, Sch{\"u}tze H (2020) Sentence meta-embeddings for unsupervised semantic textual similarity. In: Jurafsky D, Chai J, Schluter N, Tetreault J (eds) Proceedings of the 58th Annual Meeting of the Association for Computational Linguistics, pp 7027--7034. Association for Computational Linguistics, Online. doi: 10.18653/v1/2020.acl-main.628

\bibitem{b28} Guder L, Aires JP, da Silva HU, Meneguzzi F, Griebler D (2026) Sentence representations for semantic textual similarity: a systematic review. Comput Speech Lang 100:101970. doi: 10.1016/j.csl.2026.101970



\bibitem{b30} Giray L (2023) Prompt engineering with ChatGPT: a guide for academic writers. Ann Biomed Eng 51(12):2629--2633. doi: 10.1007/s10439-023-03272-4

\bibitem{b31} Liu Z, Chai Y, Li J (2024) Toward automated simulation research workflow through LLM prompt engineering design. J Chem Inf Model 65(1):114--124. doi: 10.1021/acs.jcim.4c01653

\bibitem{b32} Torkestani MS, Alameer A, Palaiahnakote S, Manosuri T (2025) Inclusive prompt engineering for large language models: a modular framework for ethical, structured, and adaptive AI. Artif Intell Rev 58(11):348. doi: 10.1007/s10462-025-11330-7

\bibitem{b33} Liu X, Wang J, Yuan X, Sun J, Dong G, Di P, Wang W, Wang D (2026) Prompting frameworks for large language models: a survey. ACM Comput Surv 58(10):255. doi: 10.1145/3789253



\bibitem{b35} Kibriya H, Khan WZ, Siddiqa A, Khan MK (2024) Privacy issues in large language models: a survey. Comput Electr Eng 120:109698. doi: 10.1016/j.compeleceng.2024.109698

\bibitem{b36} Edemacu K, Wu X (2025) Privacy preserving prompt engineering: a survey. ACM Comput Surv 57(10):255. doi: 10.1145/3729219

\bibitem{b37} Bender EM, Gebru T, McMillan-Major A, Shmitchell S (2021) On the dangers of stochastic parrots: can language models be too big? In: Proceedings of the 2021 ACM Conference on Fairness, Accountability, and Transparency (FAccT '21), pp 610--623. Association for Computing Machinery, New York. doi: 10.1145/3442188.3445922

%
\end{thebibliography}
